\documentclass{llncs}
\usepackage{bm}
\usepackage[utf8]{inputenc}
\usepackage{url}
\usepackage{amssymb}
\usepackage{amsfonts}
\usepackage{graphicx}
\usepackage{url}
\usepackage{booktabs}

\usepackage{float}
\restylefloat{figure}

\usepackage{hyperref}
\usepackage{multirow}
\usepackage{fancyvrb}
\usepackage{tikz,pgfplots}

\title{First Experiments with Neural Translation of Informal to Formal Mathematics}

\newcommand*\samethanks[1][\value{footnote}]{\footnotemark[#1]}
\author{
Qingxiang Wang
	\inst{1,2}\thanks{Supported by ERC grant no.\ 714034 \textit{SMART}}
\and Cezary Kaliszyk
  \inst{1}\samethanks
  \orcidID{\scalebox{.85}{0000-0002-8273-6059}}
\and Josef Urban
	\inst{2}\thanks{Supported by the \textit{AI4REASON} ERC Consolidator grant number 649043, and by the Czech project AI\&Reasoning CZ.02.1.01/0.0/0.0/15\_003/0000466 and the European Regional Development Fund.}}
\institute{
University of Innsbruck
\and Czech Technical University in Prague
}

\begin{document}

\maketitle{}

\begin{abstract}
  We report on our experiments to train deep neural networks
  that automatically translate informalized \LaTeX{}-written Mizar
  texts into the formal Mizar language. To the best of our knowledge, this is the first time when neural networks have been adopted in the formalization of mathematics.
  Using Luong et al.'s neural machine translation model (NMT), we tested our aligned informal-formal corpora against various hyperparameters and evaluated their results.
  Our experiments show that our best performing model configurations are able to generate correct Mizar statements on 65.73\% of the inference data, 
with the union of all models covering 79.17\%.
These results indicate that formalization through artificial neural network is a promising approach for automated formalization of mathematics.
  We present several case studies to illustrate our results.
\end{abstract}

\section{Introduction: Autoformalization}
\label{intro}
In this paper we describe our experiments with training an
end-to-end translation of \LaTeX{}-written mathematical texts to a
formal and verifiable mathematical language -- in this case the Mizar
language. This is the next step in our \emph{project to automatically learn formal understanding}~\cite{KaliszykUV15,KaliszykUVG14,KaliszykUV17} of 
mathematics and exact sciences using large corpora of alignments
between informal and formal statements.  Such machine learning and
statistical translation methods can additionally integrate strong
semantic filtering methods such as type-checking and large-theory
Automated Theorem Proving (ATP)~\cite{hammers4qed,UrbanV13}.

Since there are currently no large corpora that would align many pairs
of human-written informal \LaTeX{} formulas with their corresponding
formalization, we obtain the first corpus for the experiments
presented here by \textit{informalization}~\cite{KaliszykUV15}. This
is in general a process in which a formal text is turned into (more)
informal one. In our previous work over Flyspeck and
Mizar~\cite{KaliszykUV15,KaliszykUV17} the main informalization method
was to forget which overloaded variants and types of the mathematical
symbols are used in the formal parses. Here, we additionally use a
nontrivial transformation of Mizar to \LaTeX{} that has been developed
over two decades by Grzegorz Bancerek~\cite{bancerekmizar,bancerek2006automatic} for presenting and publishing
the Mizar articles in the journal Formalized Mathematics.\footnote{\url{https://www.degruyter.com/view/j/forma}}

Previously~\cite{KaliszykUV15,KaliszykUV17}, we have built and trained
on the smaller aligned corpora custom translation systems based on
probabilistic grammars, enhanced with semantic pruning methods such as type-checking.
Here we experiment with state-of-the-art \emph{artificial neural networks}.
It has been shown recently that given enough data, neural architectures can learn to a high degree 
the syntactic correspondence between two languages
\cite{Sutskever:2014:SSL:2969033.2969173}.
We are interested to see to what extent the neural methods can achieve meaningful translation
by training on aligned informal-formal pairs of mathematical statements.
The neural machine translation (NMT) architecture that we use
is Luong et al.'s implementation~\cite{luong17} of the sequence-to-sequence (seq2seq) model.

We will start explaining our ideas by first providing a self-contained
introduction to neural translation and the seq2seq model in
Section~\ref{Neural}.  Section~\ref{Dataset} explains how the large
corpus of aligned Mizar-\LaTeX{} formulas is created.
Section~\ref{Applying} discusses preprocessing steps and application
of NMT to our data, and Section~\ref{Evaluation} provides an
exhaustive evaluation of the neural methods available in NMT. Our main
result is that when trained on about 1 million aligned Mizar-\LaTeX{} pairs, the best
method achieves perfect translation on 65.73\% of about 100 thousand testing pairs.
Section~\ref{Conclusion} concludes and discusses the research directions opened by this work.

\section{Neural Translation}
\label{Neural}
Function approximation through artificial neural network has existed in the literature since 1940s~\cite{Goodfellow-et-al-2016}.
Theoretical results in late 80-90s have shown that it is possible to approximate an arbitrary measurable function by layers of compositions of linear and nonlinear mappings, with the nonlinear mappings satisfying certain mild properties~\cite{Cybenko1989,Hornik:1991:ACM:109691.109700}.
However, before 2010s due to limitation of computational power and lack of large training datasets, neural networks generally did not perform as well as alternative methods.

Situation changed in early 2010s when the first GPU-trained convolutional neural network outperformed all rival methods in an image classification contest~\cite{NIPS2012_4824}, in which a large labeled image dataset was used as training data.
Since then we have witnessed an enormous amount of successful applications of neural networks, culminating in 2016 when a professional Go player was defeated by a neural network-enabled Go-playing system~\cite{pre-alphago}.

Over the years many variants of neural network architectures have been
invented, and easy-to-use neural frameworks have
been built. We are particularly interested in
the sequence-to-sequence (seq2seq)
architectures~\cite{Sutskever:2014:SSL:2969033.2969173,cho-al-emnlp14} which
have achieved tremendous successes in natural language translation as
well as related tasks.  In particular, we have chosen Luong's NMT
framework~\cite{luong17} that encapsulates the Tensorflow
API gracefully and the
hyperparameters of the seq2seq model are clearly exposed at
command-line level.  This allows us to quickly and systematically
experiment with our data.

\subsection{The Seq2seq Model}

\begin{figure}
	\centering
	\includegraphics[width=0.5\linewidth]{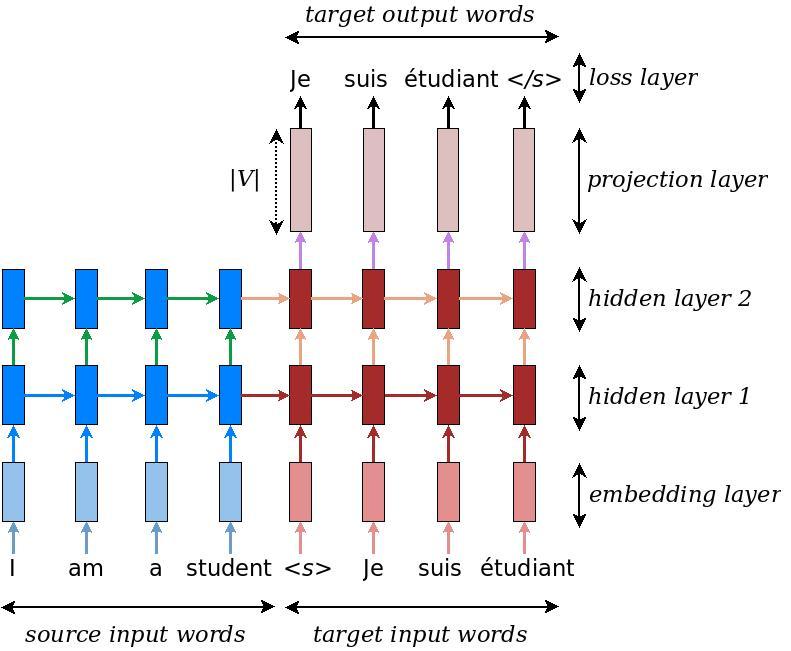}
	\caption{Seq2seq model (adapted from Luong et al.~\cite{luong17})}
	\label{fig:seq2seq}
\end{figure}

A seq2seq model is a network which consists of an encoder and a decoder (the left and  right part in Fig.~\ref{fig:seq2seq}).
During training, the encoder takes in a sentence one word at a time from the source language, and the decoder takes in the corresponding sentence from the target language.
The network generates another target sentence and a loss function is computed based on the input target sentence and the generated target sentence.
As each word in a sentence will be embedded into the network as a real vector, the whole network can be considered as a complicated function from a finite-dimensional real vector space to the reals.
Training of the neural network amounts to conducting optimization based on this function.

\begin{figure}
	\centering
	\includegraphics[width=0.4\linewidth]{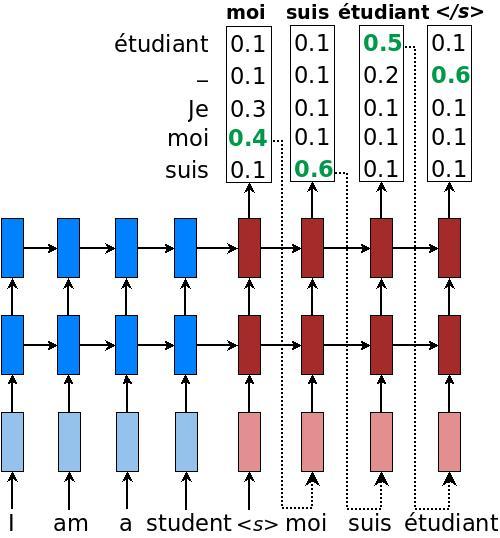}
	\caption{Inference of seq2seq model (adapted from Luong et al.~\cite{luong17})}
	\label{fig:greedy_dec}
\end{figure}

When the training is complete, the neural network can be used to generate translations by \emph{inferring} from (translating of) 
unseen source sentences.
During inference, only the source sentence is provided.
A target sentence is then generated word after word from the decoder by conducting greedy evaluation with the probabilistic model represented by the trained neural network (Fig.~\ref{fig:greedy_dec}).

\subsection{RNN and the RNN Memory Cell}\label{ssec:rnn}

\begin{figure}
	\centering
	\includegraphics[width=0.7\linewidth]{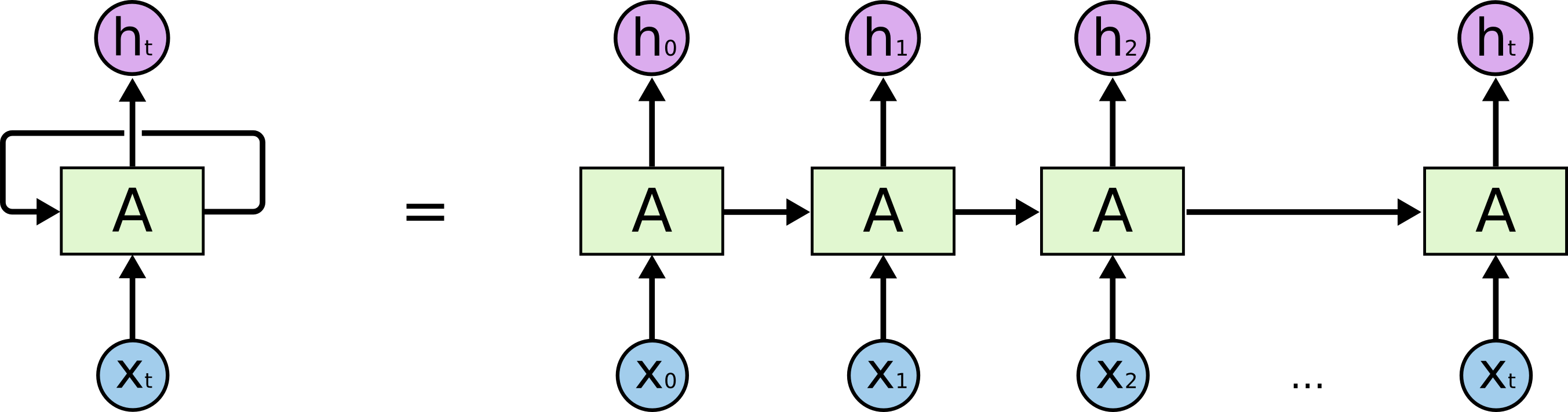}
	\caption{RNN cell and its rolled-out format (adapted from Olah's blog~\cite{colah-lstm})}
	\label{fig:RNN-unrolled}
\end{figure}

The architectures of the encoder and the decoder inside the seq2seq model are similar, each of which consists of multiple layers of \emph{recurrent neural networks} (RNNs).
A typical RNN consists of one memory cell, which takes input word tokens (in vector format) and updates its parameters iteratively.
An RNN cell is typically presented in literature in the \emph{rolled-out format}  (Fig.~\ref{fig:RNN-unrolled}), though the same memory cell is used and the same set of parameters are being updated during training.

\begin{figure}
	\centering
	\includegraphics[width=0.7\linewidth]{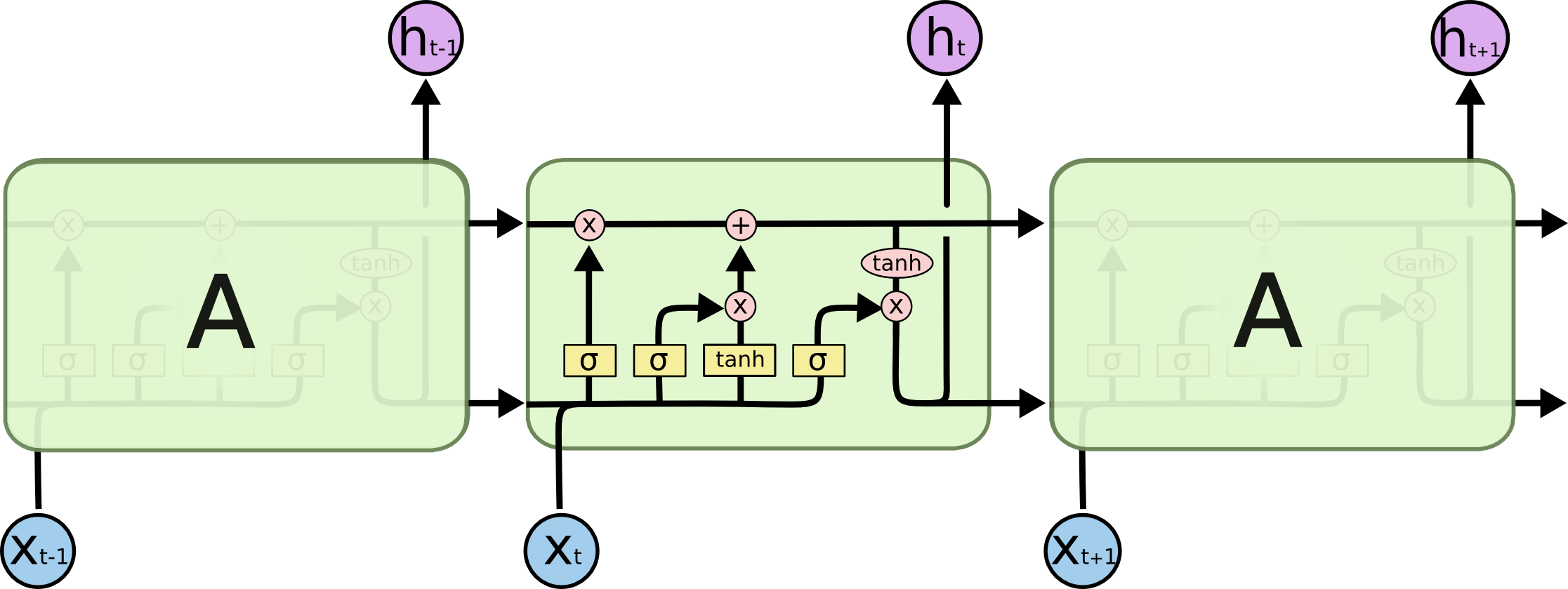}
	\caption{Close-up look of an LSTM cell (adapted from Olah's blog~\cite{colah-lstm})}
	\label{fig:LSTM3-chain}
\end{figure}

Inside each memory cell there is an intertwined combination of linear and nonlinear transformations (Fig.~\ref{fig:LSTM3-chain}).
These transformations are carefully chosen to mimic the cognitive process of keeping, retaining and forgetting information.

\begin{figure}
	\centering
	\includegraphics[width=0.4\linewidth]{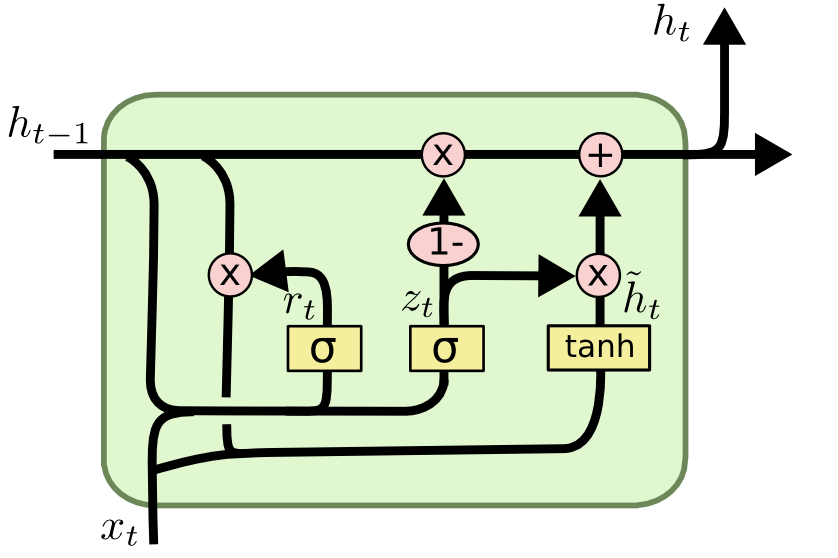}
	\caption{Gated recurrent unit (adapted from Olah's blog~\cite{colah-lstm})}
	\label{fig:LSTM3-var-GRU}
\end{figure}

Only differentiable functions are used to compose the memory cell, so the overall computation is also a differentiable function and gradient-based optimization can be adopted.
In addition, the computation is designed in so that the derivative of the memory cell's output with respect to its input is always close to one.
This ensures that the problem of vanishing or exploding gradients is avoided when conducting differentiation using the chain rule.
Several variants of memory cells exist.
The most common are the \emph{long short-term memory} (LSTM) in Fig.~\ref{fig:LSTM3-chain} and the \emph{gated recurrent unit} (GRU), in Fig.~\ref{fig:LSTM3-var-GRU}.

\subsection{Attention Mechanism}

\begin{figure}
	\centering
	\includegraphics[width=0.5\linewidth]{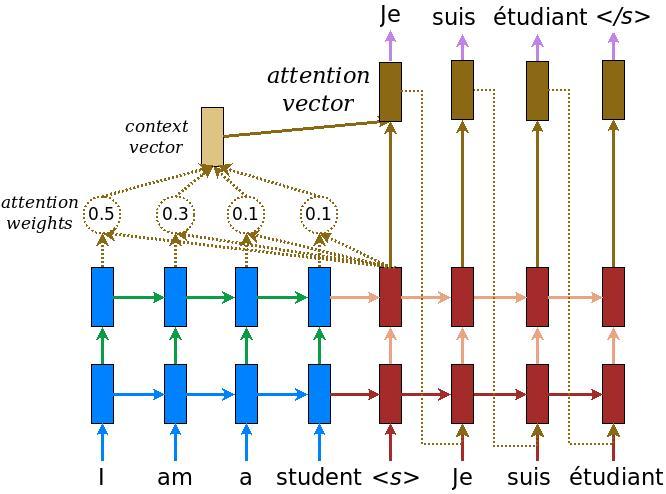}
	\caption{Neural network with attention mechanism (adapted from Luong et al.~\cite{luong17})}
	\label{fig:attention_mechanism}
\end{figure}

The current seq2seq model has a limitation: in the first iteration the decoder obtains all the information from the encoder, which is unnecessary as not all parts of the source sentence contribute equally to particular parts of the target sentence.
The \emph{attention} mechanism is used to overcome this limitation by adding special layers in parallel to the decoder (Fig.~\ref{fig:attention_mechanism}).
These special layers compute scores which can provide a weighting mechanism to let the decoder decide how much emphasis should be put on certain parts of the source sentence when translating a certain part of the target sentence.
There are also several variants of the attention mechanism, depending on how the scores are computed or how the input and output are used. 
In our experiments, we will explore all the attention mechanisms provided by the NMT framework and evaluate their performance on the Mizar-\LaTeX{} dataset.

\section{The Informalized Dataset}
\label{Dataset}
State-of-the-art neural translation methods generally require large
corpora consisting of many pairs of aligned sentences (e.g. in German
and English). The lack of aligned data in our case has been a bottleneck preventing experiments
with end-to-end neural translation from informal to formal mathematics.  The
approach that we have used so far for experimenting with non-neural
translation methods is to take a large formal corpus such as
Flyspeck~\cite{HalesABDHHKMMNNNOPRSTTTUVZ15} or
Mizar~\cite{BancerekBGKMNPU15} and apply various
\textit{informalization}
(\textit{ambiguation})~\cite{KaliszykUV15,KaliszykUV17}
transformations to the formal sentences to obtain their less formal
counterparts.  Such transformations include e.g. forgetting which
overloaded variants and types of the mathematical symbols are used in
the formal parses, forgetting of explicit casting functors,
bracketing, etc. These transformations result in more human-like and
ambiguous sentences that (in particular in the case of Mizar) resemble the
natural language style input to ITP systems, but the sentences
typically do not include more complicated symbol transformations that occur naturally in \LaTeX{}.

There are several formalizations such as Flyspeck, the Coq proof of
the Odd-Order theorem, the Mizar formalization of the Compendium of
Continuous Lattices (CCL) that come with a high-level alignment of the
main theorems in the corresponding (\LaTeX{}-written) books to the main formalized
theorems. However, such mappings are so far quite sparse: e.g., there are about
500 alignments between Flyspeck and its informal
book~\cite{hales-dense}.  Instead, we have decided to obtain the first
larger corpus of aligned \LaTeX{}/formal sentences again by
informalization.  Our requirement is that the informalization should
be nontrivial, i.e., it should target a reasonably rich subset of
\LaTeX{} and the transformations should go beyond simple symbol
replacements. 

The choice that we eventually made is to use the Mizar
translation to \LaTeX{}. This translation has been developed for more than two
decades by the Mizar team~\cite{TR89} and specifically by Grzegorz
Bancerek~\cite{bancerekmizar,bancerek2006automatic} for presenting and
publishing the Mizar articles in the journal Formal Mathematics. 
This translation is relatively
nontrivial~\cite{bancerek2006automatic}. It starts with 
user-defined translation patterns for different basic objects of the
Mizar logic: functors, predicates, and type constructors such as
adjectives, modes and structures. Quite complicated mechanisms
are also used to decide on the use of brackets, the uses of singular/plural cases,
regrouping of conjunctive formulas, etc. Tables~\ref{tab:xboole_1} and \ref{tab:bhsp_2} show 
examples of this translation for theorems
\texttt{XBOOLE\_1:1}\footnote{\url{http://grid01.ciirc.cvut.cz/~mptp/7.13.01_4.181.1147/html/xboole_1\#T1}} and
\texttt{BHSP\_2:3}\footnote{\url{http://grid01.ciirc.cvut.cz/~mptp/7.13.01_4.181.1147/html/bhsp_2\#T13}}, together
with their tokenized form used for the neural training and translation.
\begin{table}[htbp!]
\def\arraystretch{1.3}%
\begin{tabular}{p{0.195\linewidth}p{0.0\linewidth}p{0.78\linewidth}}
\toprule
Rendered \LaTeX{} & & If $X \subseteq Y \subseteq Z$, then $X \subseteq Z$. \\
Mizar & &
\begin{BVerbatim}
X c= Y & Y c= Z implies X c= Z;
\end{BVerbatim}
\\
Tokenized Mizar & &
\begin{BVerbatim}
X c= Y & Y c= Z implies X c= Z ;
\end{BVerbatim}
\\
\LaTeX{} & &
\begin{BVerbatim}
If $X \subseteq Y \subseteq Z$, then $X \subseteq Z$.
\end{BVerbatim}
\\
Tokenized \LaTeX{} & &
\begin{BVerbatim}
If $ X \subseteq Y \subseteq Z $ , then $ X \subseteq Z $ .
\end{BVerbatim}
\\
\bottomrule
\end{tabular}
\caption{Theorem 1 in \texttt{XBOOLE\_1}}\label{tab:xboole_1}
\end{table}
\begin{table}[htbp!]
\def\arraystretch{2.6}%
\begin{tabular}{p{0.195\linewidth}p{0.0\linewidth}p{0.78\linewidth}}
\toprule
\raisebox{-2mm}{Rendered \LaTeX{}} & & Suppose $ { s_{8} } $ is convergent and $ { s_{7} } $ is convergent . 
Then $ \mathop { \rm lim } ( { s_{ 8 } } { + } { s_{7} } ) \mathrel { = } 
\mathop { \rm lim } { s_{8}} { + } \mathop { \rm lim } { s_{7} } $  \\
\raisebox{2mm}{Mizar} & &
\begin{BVerbatim}
seq1 is convergent & seq2 is convergent implies lim(seq1
+seq2)=(lim seq1)+(lim seq2);
\end{BVerbatim}
\\
\raisebox{2mm}{Tokenized Mizar} & &
\begin{BVerbatim}
seq1 is convergent & seq2 is convergent implies lim ( 
seq1 + seq2 ) = ( lim seq1 ) + ( lim seq2 ) ;
\end{BVerbatim}
\\
\raisebox{6mm}{\LaTeX{}} & &
\begin{BVerbatim}
Suppose ${s_{8}}$ is convergent and ${s_{7}}$ is 
convergent. Then $\mathop{\rm lim}({s_{8}}{+}{s_{7}})
\mathrel{=}\mathop{\rm lim}{s_{8}}{+}
\mathop{\rm lim}{s_{7}}$
\end{BVerbatim}
\\
\raisebox{8mm}{Tokenized \LaTeX{}} & & 
\begin{BVerbatim}
Suppose $ { s _ { 8 } } $ is convergent and $ { s _ { 7
} } $ is convergent . Then $ \mathop { \rm lim } ( { s 
_ { 8 } } { + } { s _ { 7 } } ) \mathrel { = } \mathop 
{ \rm lim } { s _ { 8 } } { + } \mathop { \rm lim } { s
_ { 7 } } $
\end{BVerbatim}
\\
\bottomrule
\end{tabular}
\caption{Theorem 3 in \texttt{BHSP\_2}}\label{tab:bhsp_2}
\end{table}

Since Bancerek's technology is only able to translate Mizar formal abstracts into Latex, in order to obtain the maximum amount of data, we modified the latest experimental
Mizar-to-\LaTeX{} XSL stylesheets that include the option to produce
all the proof statements. During the translation of a Mizar article we track for every
proof-internal formula its starting position (line and column) in the
corresponding Mizar article, marking the formulas with these positions
in the generated \LaTeX{} file. We then extract each formula tagged
with its position $P$ from the \LaTeX{} file, align it with the
Mizar formulas starting at position $P$, and apply further data processing to them (Section~\ref{preprocessing}). 
This results in about one million aligned pairs of \LaTeX{}/Mizar sentences.

\section{Applying Neural Translation to Mizar}
\label{Applying}

\subsection{Data Preprocessing}
\label{preprocessing}
To adapt our data to NMT, the \LaTeX{} sentences and their
corresponding Mizar sentences must be properly tokenized
(Table~\ref{tab:xboole_1} and~\ref{tab:bhsp_2}).  In addition,
distinct word tokens from both \LaTeX{} and Mizar must also be
provided as vocabulary files.

In Mizar formulas, tokens can be and often are concatenated -- as
e.g. in \texttt{n<m}.  We used each article's symbol and identifier
files produced by the Mizar accommodator and parser to separate such
tokens.  For \LaTeX{} sentences, we decided to consider dollar signs,
brackets, parentheses, carets and underscores as separate tokens. We
keep tags starting with backslash intact and leave all the font
information (e.g. romanization or emphasis). Cross-referencing tags,
styles for itemization as well as other typesetting information are
removed.

\subsection{Division of Data}

Luong's NMT model requires a small set of development data and test data in addition to training data. 
To conduct the full training-inference process the raw data needs to be divided into four parts.
Our preprocessed data contains 1,056,478 pairs of Mizar-\LaTeX{} sentences.
In order to achieve a 90:10 training-to-inference ratio we randomly divide our data into the following:
\begin{itemize}
\item 947,231 pairs of sentences of training data.
\item 2,000 pairs of development data (for NMT model selection).
\item 2,000 pairs of test data (for NMT model evaluation).
\item 105,247 pairs of inference (testing) data.
\item 7,820 and 16,793 unique word tokens generated for the vocabulary files of \LaTeX{} and Mizar sentences, respectively.
\end{itemize}

For our partition, there are 57,145 lines of common latex sentences in both the training set and the inference set, making up to 54.3\% of the inference set.
This is expected as mathematical proofs involve a lot of common basic proof steps. Therefore, in addition to correct translations, we are also interested in correct translations in the 48,102 non-overlapping sentences.

\subsection{Choosing Hyperparameters}
Luong's NMT model provides around 70 configurable hyperparameters, many of which can affect the architecture of the neural network and in turn affect the training results.
In our experiments, we decided to evaluate our model with respect to the following 7 hyperparameters that are the most relevant to the behavior of the seq2seq model (Table~\ref{tab:hyper}), while keeping other hyperparameters (those that are more auxiliary, experimental or non-recommended for change) at their default.
Selected common hyperparameters are listed in Table~\ref{tab:common}.
\begin{table}[htbp!]
\begin{tabular}{p{0.49\linewidth}p{0.49\linewidth}}
\toprule
Name & Default Value \\
\midrule
Number of training steps & 12,000 \\
Learning rate & 1.0 (0.001 when using Adam optimizer) \\
Forget bias for LSTM cell & 1.0 \\
Dropout rate & 0.2 \\
Batch size & 128 \\
Decoding type & greedy \\
\bottomrule
\end{tabular}
\caption{Common network hyperparameters across experiments}\label{tab:common}
\end{table}
\begin{table}[htbp!]
\begin{tabular}{p{0.15\linewidth}p{0.55\linewidth}p{0.275\linewidth}}
\toprule
Name & Description & Value \\
\midrule
unit type & Type of the memory cell in RNN & LSTM (default)\newline GRU\newline Layer-norm LSTM \\
attention & The attention mechanism & No Attention (default)\newline (Normed) Bahdanau\newline (Scaled) Luong\\
nr. of layers & RNN layers in encoder and decoder & 2 layers (default)\newline 
3/4/5/6 layers\\
residual & Enables residual layers (to overcome exploding/vanishing gradients) & False (default)\newline True \\
optimizer & The gradient-based optimization method & SGD (default)\newline Adam \\
encoder type & Type of encoding methods for input sentences & Unidirectional (default)\newline Bidirectional \\
nr. of units & The dimension of parameters in a memory cell & 128 (default)\newline 
256/512/1024/2048 \\
\bottomrule
\end{tabular}
\caption{Hyperparameters for seq2seq model}\label{tab:hyper}
\end{table}

\section{Evaluation}
\label{Evaluation}
The results are evaluated by four different metrics: 1) perplexity; 2) the BLEU rate of the final test data set; 3) the number and percentage of identical statements within all the 105,247 inference sentences and 4) the number and percentage of identical statements within the 48,102 non-overlapping inference sentences.
Perplexity measures the difficulty of generating correct words in a sentence, and the BLEU rate gives a score on the quality of the overall translation.
Details explaining perplexity and the BLEU rate can be found in~\cite{DBLP:journals/corr/Neubig17} and~\cite{Papineni02bleu:a}, respectively.
Due to the abundance of hyperparameters, we decided to do our experiments progressively, by first comparing a few basic hyperparameters, fixing the best choices and then comparing the other hyperparameters.
The basic hyperparameters we chose are the type of memory cell and the attention mechanism.

\subsection{Choosing the Best Memory Cell and Attention Mechanism}
From Table~\ref{tab:cell} we can see that GRU and LSTM perform similarly
and both perform better than Layer-normed LSTM.  As LSTM
performed slightly better than GRU we fixed our memory cell to be LSTM
for further experiments.
\footnote{Since training and inference involve randomness, the final results are not identical across trials, though our experience showed that the variation of the inference metrics are small.}

Published NMT evaluations show that the attention mechanism results in better performance in translation tasks. 
Our experiments confirm this fact and also show that the Normed Bahdanu attention, Luong attention and Scaled Luong attention are better than Bahdanau attention (Table~\ref{tab:attention}).
Among them we picked the best-performing Scaled Luong attention as our new default and used this attention for our further experiments.
\begin{table}
\begin{tabular}{p{0.25\linewidth}p{0.135\linewidth}p{0.135\linewidth}p{0.225\linewidth}p{0.21\linewidth}}
\toprule
Parameter & Final Test\newline Perplexity & Final Test\newline BLEU & Identical\newline Statements (\%) & Identical\newline No-overlap (\%)\\
\midrule
LSTM & \textbf{3.06} & \textbf{41.1} & \textbf{40121 (38.12\%)} & \textbf{6458 (13.43\%)} \\
GRU & 3.39 & 34.7 & 37758 (35.88\%) & 5566 (11.57\%)\\
Layer-norm LSTM & 11.35 & 0.4 & 11200 (10.64\%) & 1 (0\%) \\
\bottomrule
\end{tabular}
\caption{Evaluation on type of memory cell (attention not enabled)}\label{tab:cell}
\end{table}

\begin{table}
\begin{tabular}{p{0.25\linewidth}p{0.135\linewidth}p{0.135\linewidth}p{0.225\linewidth}p{0.21\linewidth}}
\toprule
Parameter & Final Test\newline Perplexity & Final Test\newline BLEU & Identical\newline Statements (\%) & Identical\newline No-overlap (\%) \\
\midrule
No Attention & 3.06 & 41.1 & 40121 (38.12\%) & 6458 (13.43\%)\\
Bahdanau & 3 & 40.9 & 44218 (42.01\%) & 8440 (17.55\%)\\
Normed Bahdanau & 1.92 & 63.5 & 60192 (57.19\%) & 18057 (37.54\%)\\
Luong & \textbf{1.89} & 64.8 & 60151 (57.15\%) & 18013 (37.45\%)\\
Scaled Luong & 2.13 & \textbf{65} & \textbf{60703 (57.68\%)} & \textbf{18105 (37.64\%)}\\
\bottomrule
\end{tabular}
\caption{Evaluation on type of attention mechanism (LSTM cell)}\label{tab:attention}
\end{table}

\subsection{The Effect of Optimizers, Residuals and Encodings with respect to Layers}
After fixing the memory cell and the attention mechanism, 
we tried the effects of the optimizer types
and of the encoding mechanisms on our data with respect to the number of
the RNN layers. We also experiment with enabling the residual layers.  The
results are shown in Table~\ref{tab:layers1}. We can observe that:
\begin{enumerate}
\item For RNN because of the vanishing gradient problem the result
  generally deteriorates when the number of layers becomes higher.  Our
  experiments confirm this: the best-performing architecture has 3-layers.
\item Residuals can be used to alleviate the effect of vanishing
  gradients.  We see from Table~\ref{tab:layers1} that the results are
  generally better with residual layers enabled, though there are
  cases when residuals produce failures in training.
\item The NaN values are caused by the overflow of the optimization metric (bleu rate).
For some hyperparameter combinations, it happens that the metric will get
worse as training progresses, which ultimately leads to overflow and subsequent early stop of the training phase.
Our experiments show that this overflow reappears with respect to multiple times of trainings.
\item It is interesting that the Adam optimizer, bidirectional
  encoding and combinations of them can also alleviate the effect of
  vanishing gradients.
\item  The Adam optimizer performs
  generally better than the SGD optimizer and Bidirectional encoding performs
  better with less layers.\footnote{Bidirectional
    encoding only works on even number of layers.}
\item The number of layers seems to matter less in our model than other
  parameters such as optimizers and encoding mechanisms, though it is
  notable that the more layers the longer the training time.
\item The number of identical non-overlapping statements is generally proportional to the total number of identical statements.
\end{enumerate}
\begin{table}[htbp!]
\begin{tabular}{p{0.28\linewidth}p{0.135\linewidth}p{0.135\linewidth}p{0.225\linewidth}p{0.21\linewidth}}
\toprule
Parameter & Final Test\newline Perplexity & Final Test\newline BLEU & Identical\newline Statements (\%) & Identical\newline No-overlap (\%) \\
\midrule
2-Layer & 3.06 & 41.1 & 40121 (38.12\%) & 6458 (13.43\%) \\

3-Layer & 2.10 & 64.2 & 57413 (54.55\% & 16318 (33.92\%) \\

4-Layer & 2.39 & 45.2 & 49548 (47.08\%) & 11939 (24.82\%)\\

5-Layer & 5.92 & 12.8 & 29207 (27.75\%) & 2698 (5.61\%)\\

6-Layer & 4.96 & 20.5 & 29361 (27.9\%) & 2872 (5.97\%)\\

2-Layer Residual & 1.92 & 54.2 & 57843 (54.96\%) & 16511 (34.32\%)\\

3-Layer Residual & 1.94 & 62.6 & 59204 (56.25\%) & 17396 (36.16\%)\\

4-Layer Residual & 1.85 & 56.1 & 59773 (56.79\%) & 17626 (36.64\%)\\

5-Layer Residual & 2.01 & 63.1 & 59259 (56.30\%) & 17327 (36.02\%)\\

6-Layer Residual & NaN & 0 & 0 (0\%) & 0 (0\%)\\

2-Layer Adam & 1.78 & 56.6 & 61524 (58.46\%) & 18635 (38.74\%)\\

3-Layer Adam & 1.91 & 60.8 & 59005 (56.06\%) & 17213 (35.78\%)\\

4-Layer Adam & 1.99 & 51.8 & 57479 (54.61\%) & 16288 (33.86\%)\\

5-Layer Adam & 2.16 & 54.3 & 54670 (51.94\%) & 14769 (30.70\%)\\

6-Layer Adam & 2.82 & 37.4 & 46555 (44.23\%) & 10196 (21.20\%)\\

2-Layer Adam Res. & 1.75 & 56.1 & 63242 (60.09\%) & 19716 (40.97\%)\\

3-Layer Adam Res. & 1.70 & 55.4 & 64512 (61.30\%) & 20534 (42.69\%)\\

4-Layer Adam Res. & 1.68 & 57.8 & 64399 (61.19\%) & 20353 (42.31\%)\\

5-Layer Adam Res. & 1.65 & 64.3 & 64722 (61.50\%) & 20627 (42.88\%)\\

6-Layer Adam Res. & 1.66 & 59.7 & 65143 (61.90\%) & 20854 (43.35\%)\\

2-Layer Bidirectional & 2.39 & \textbf{69.5} & 63075 (59.93\%) & 19553 (40.65\%)\\

4-Layer Bidirectional & 6.03 & 63.4 & 58603 (55.68\%) & 17222 (35.80\%)\\

6-Layer Bidirectional & 2 & 56.3 & 57896 (55.01\%) & 16817 (34.96\%)\\

2-Layer Adam Bi. & 1.84 & 56.9 & 64918 (61.68\%) & 20830 (43.30\%)\\

4-Layer Adam Bi. & 1.94 & 58.4 & 64054 (60.86\%) & 20310 (42.22\%)\\

6-Layer Adam Bi. & 2.15 & 55.4 & 60616 (57.59\%) & 18196 (37.83\%)\\

2-Layer Bi. Res. & 2.38 & 24.1 & 47531 (45.16\%) & 11282 (23.45\%)\\

4-Layer Bi. Res. & NaN & 0 & 0 (0\%) & 0 (0\%)\\

6-Layer Bi. Res. & NaN & 0 & 0 (0\%) & 0 (0\%)\\

2-Layer Adam Bi. Res. & 1.67 & 62.2 & 65944 (62.66\%) & 21342 (44.37\%)\\

4-Layer Adam Bi. Res. & \textbf{1.62} & 66.5 & 65992 (62.70\%) & 21366 (44.42\%)\\

6-Layer Adam Bi. Res. & 1.63 & 58.3 & \textbf{66237 (62.93\%)} & \textbf{21404 (44.50\%)} \\
\bottomrule
\end{tabular}
\caption{Evaluation on various hyperparameters w.r.t. layers}\label{tab:layers1}
\end{table}

\subsection{The Effect of the Number of Units and the Final Result}
We now train our models by fixing other hyperparameters and variating the number of units.
Our results in
Table~\ref{tab:units} show that performance generally gets better
until 1024 units. The performance decreases when the number of units
reaches 2048, which might indicate that the model starts to
overfit. We have so far only used CPU versions of Tensorflow.
The training real times in hours of our multi-core Xeon
E5-2690 v4 2.60GHz servers with 28 hyperthreading cores
are also included to illustrate the usage of computational resources
with respect to the number of units.

The best result achieved with 1024 units shows that after training for
11 hours on the corpus of the 947231 aligned Mizar-\LaTeX{} pairs, we
can automatically translate with perfect accuracy 69179 (65.73\%)
of the 105247 testing pairs. Given that the translation includes quite
nontrivial transformations, this is a surprisingly good
performance. Also, by manually inspecting the remaining
misclassifications we have found that many of those are actually
semantically correct translations, typically choosing different but
synonymous expressions. A simple example of such synonyms is the Mizar
expression \texttt{for x st P(x) holds Q(x)}, which can be
alternatively written as \texttt{for x holds P(x) implies Q(x)}. Since
there are many such synonyms on various levels and they are often
context-dependent, the true \emph{semantic performance} of the
translator will have to be measured by further applying the
translation~\cite{Urban06} from Mizar to MPTP/TPTP to the current results, and
calling ATP systems to establish equivalence with the original Mizar
formula as we do for Flyspeck in~\cite{KaliszykUV17}. This is left as
future work.
\begin{table}[htbp!]
\begin{tabular}{p{0.145\linewidth}p{0.13\linewidth}p{0.135\linewidth}p{0.22\linewidth}p{0.215\linewidth}p{0.135\linewidth}}
\toprule
Parameter & Final Test\newline Perplexity & Final Test\newline BLEU & Identical\newline Statements (\%) & Identical\newline No-overlap (\%) & Training\newline Time (hrs.)\\
\midrule
128 Units & 3.06 & 41.1 & 40121 (38.12\%) & 6458 (13.43\%) & 1\\

256 Units & 1.59 & 64.2 & 63433 (60.27\%) & 19685 (40.92\%) & 3\\

512 Units & 1.6 & \textbf{67.9} & 66361 (63.05\%) & 21506 (44.71\%) & 5\\

1024 Units & \textbf{1.51} & 61.6 & \textbf{69179 (65.73\%)} & \textbf{22978 (47.77\%)} & 11\\

2048 Units & 2.02 & 60 & 59637 (56.66\%) & 16284 (33.85\%) & 31\\
\bottomrule
\end{tabular}
\caption{Evaluation on number of units}\label{tab:units}
\end{table}

\subsection{Greedy Covers and Edit Distances}
We illustrate the combined performance of translation by comparing against selected collections of models. 
In Table~\ref{tab:distances} "Top-$n$ Greedy Cover" denotes a list of $n$ models such that each model in the list gives the maximum increase of correct translations from the previous model.
In addition, we also measure the percentage of sentences (both overlap and no-overlap part) that are nearly correct.
The metric of nearness we use is the word-level minimum editing distance (Levenshtein distance).
We can see from Table~\ref{tab:distances} that reasonably correct translations can be generated by just using a combination of a few models.

\begin{table}[htbp!]
\begin{tabular}{p{0.30\linewidth}p{0.25\linewidth}p{0.1\linewidth}p{0.1\linewidth}p{0.1\linewidth}p{0.1\linewidth}}
\toprule
 & Identical\newline Statements & 0 & $\leq$ 1 & $\leq$ 2 & $\leq$ 3 \\
\midrule

Best Model \newline \tiny{- 1024 Units} & 69179 (total)\newline 22978 (no-overlap) & 65.73\%\newline 47.77\% & 74.58\%\newline 59.91\% & 86.07\%\newline 70.26\% & 88.73\%\newline 74.33\% \\

& & & & & \\
Top-5 Greedy Cover \newline \tiny{- 1024 Units} \newline \tiny{- 4-Layer Bi. Res.} \newline \tiny{- 512 Units} \newline \tiny{- 6-Layer Adam Bi. Res.} \newline \tiny{- 2048 Units} & 78411 (total)\newline 28708 (no-overlap) & 74.50\%\newline 59.68\% & 82.07\%\newline 70.85\% & 87.27\%\newline 78.84\% & 89.06\%\newline 81.76\% \\

& & & & & \\
Top-10 Greedy Cover \newline \tiny{- 1024 Units} \newline \tiny{- 4-Layer Bi. Res.} \newline \tiny{- 512 Units} \newline \tiny{- 6-Layer Adam Bi. Res.} \newline \tiny{- 2048 Units} \newline \tiny{- 2-Layer Adam Bi. Res.} \newline \tiny{- 256 Units} \newline \tiny{- 5-Layer Adam Res.} \newline \tiny{- 6-Layer Adam Res.} \newline \tiny{- 2-Layer Bi. Res.} & 80922 (total)\newline 30426 (no-overlap) & 76.89\%\newline 63.25\% & 83.91\%\newline 73.74\% & 88.60\%\newline 81.07\% & 90.24\%\newline 83.68\% \\

& & & & & \\
Union of All 39 Models & 83321 (total)\newline 32083 (no-overlap) & 79.17\%\newline 66.70\% & 85.57\%\newline 76.39\% & 89.73\%\newline 82.88\% & 91.25\%\newline 85.30\% \\
\bottomrule
\end{tabular}
\caption{Coverage w.r.t. a set of models and edit distances}\label{tab:distances}
\end{table}

\subsection{Translating from Mizar to \LaTeX{}}
It is interesting to see how the seq2seq model performs on our
data when we treat Mizar as the source language and \LaTeX{} as
the target language, thus emulating Bancerek's translation toolchain. 
The results in Table~\ref{tab:rev} show that the
model is still able to achieve meaningful translations from Mizar to
\LaTeX{}, though the translation quality is generally not yet as good as
in the other direction.
\begin{table}
\begin{tabular}{p{0.35\linewidth}p{0.15\linewidth}p{0.15\linewidth}p{0.15\linewidth}p{0.15\linewidth}}
\toprule
Parameter & Final Test\newline Perplexity & Final Test\newline BLEU & Identical\newline Statements & Percentage \\
\midrule
512 Units Bidirectional \newline Scaled Luong & 2.91 & 57 & 54320 & 51.61\% \\
\bottomrule
\end{tabular}
\caption{Evaluation on number of units}\label{tab:rev}
\end{table}

\section{A Translation Example}
To illustrate the training of the neural network, we pick a specific
example (again \texttt{BHSP\_2:3} as in Section~\ref{Dataset}) and
watch how the translation changes as the training progresses.  We can
see from Table~\ref{tab:snapshot}
that the model produces mostly gibberish in the
early phases of the training.  As the training progresses, the
generated sentence starts to look more like the correct Mizar
statement.  It is interesting to see that the neural network is
able to learn the matching of parentheses and correct labeling of
identifiers.
\begin{table}
\def\arraystretch{2.5}%
\begin{tabular}{lp{0.01\linewidth}p{0.79\linewidth}}
\toprule
\raisebox{-2mm}{Rendered \LaTeX{}} & & Suppose $ { s_{8} } $ is convergent and $ { s_{7} } $ is convergent . 
Then $ \mathop { \rm lim } ( { s_{ 8 } } { + } { s_{7} } ) \mathrel { = } 
\mathop { \rm lim } { s_{8}} { + } \mathop { \rm lim } { s_{7} } $  \\
\raisebox{6mm}{Input \LaTeX{}} & &
\begin{BVerbatim}
Suppose $ { s _ { 8 } } $ is convergent and $ { s _ { 7 } } 
$ is convergent . Then $ \mathop { \rm lim } ( { s _ { 8 } 
} { + } { s _ { 7 } } ) \mathrel { = } \mathop { \rm lim } 
{ s _ { 8 } } { + } \mathop { \rm lim } { s  _ { 7 } } $ .
\end{BVerbatim}
\\
\raisebox{2mm}{Correct} & &
\begin{BVerbatim}
seq1 is convergent & seq2 is convergent implies lim ( seq1
+ seq2 ) = ( lim seq1 ) + ( lim seq2 ) ;
\end{BVerbatim}
\\
\raisebox{2mm}{Snapshot-1000} & &
\begin{BVerbatim}
x in dom f implies ( x * y ) * ( f | ( x | ( y | ( y | y )
) ) ) = ( x | ( y | ( y | ( y | y ) ) ) ) ) ;
\end{BVerbatim}
\\[-3mm]
Snapshot-3000 & &\verb!seq is convergent & lim seq = 0c implies seq = seq ;!\\
\raisebox{2mm}{Snapshot-5000} & &
\begin{BVerbatim}
seq1 is convergent & lim seq2 = lim seq2 implies lim_inf 
seq1 = lim_inf seq2 ;
\end{BVerbatim}
\\
\raisebox{2mm}{Snapshot-7000} & &
\begin{BVerbatim}
seq is convergent & seq9 is convergent implies 
lim ( seq + seq9 ) = ( lim seq ) + ( lim seq9 ) ;
\end{BVerbatim}
\\
\raisebox{2mm}{Snapshot-9000} & &
\begin{BVerbatim}
seq1 is convergent & lim seq1 = lim seq2 implies ( seq1 
+ seq2 ) + ( lim seq1 ) = ( lim seq1 ) + ( lim seq2 ) ;
\end{BVerbatim}
\\
\raisebox{2mm}{Snapshot-12000} & &
\begin{BVerbatim}
seq1 is convergent & seq2 is convergent implies
lim ( seq1 + seq2 ) = ( lim seq1 ) + ( lim seq2 ) ;
\end{BVerbatim}
\\
\bottomrule
\end{tabular}
\caption{Translation with respect to training steps}\label{tab:snapshot}
\end{table}

\section{Conclusion and Future Work}
\label{Conclusion}
We for the first time harnessed neural networks in the formalization of mathematics.  Due to the lack of aligned informal-formal
corpora, we generated informalized \LaTeX{} from Mizar by using and modifying
the current translation done for the journal Formalized
Mathematics.  Our results show that for a significant proportion of
the inference data, neural network is able to generate correct Mizar
statements from \LaTeX{}. 
In particular, when trained on the 947,231 aligned Mizar-\LaTeX{} pairs, the best method achieves perfect translation on 65.73\% of the 105,247 test pairs, and the union of all methods produces perfect translations on 79.17\% of the test pairs.

Even though these are results on a synthetic dataset, such a good
performance is surprising to us and also very encouraging. It means
that state-of-the-art neural methods are capable of learning quite
nontrivial informal-to-formal transformations, and have a great
potential to help with automating computer understanding of
mathematical and scientific writings.

It is also clear that many of the translations that are currently
classified by us as imperfect (i.e., syntactically different from the
aligned formal statement) are semantically correct. This is due to
a number of synonymous formulations allowed by the Mizar language.
Obvious future work thus includes a full semantic evaluation, i.e.,
using translation to MPTP/TPTP and ATP systems to check if the
resulting formal statements are equivalent to their aligned
counterparts. As in~\cite{KaliszykUV15,KaliszykUV17}, this will likely
also show that the translator can produce semantically different, but
still provable statements and conjectures.

Another line of research opened by these results is an
extension of the translation to full informalized Mizar proofs, then
to the ProofWiki corpus aligned by Bancerek recently to Mizar, and
(using these as bridges) eventually to arbitrary \LaTeX{} texts.
The power and the limits of the current neural architectures in automated
formalization and reasoning is worth of further understanding, and we
are also open to the possibility of adapting existing formalized
libraries to tolerate the great variety of natural language proofs.

\bibliographystyle{abbrv}
\bibliography{ate11}

\section{Appendix A: Effect of Training Steps}
\begin{table}
\def\arraystretch{2.3}%
\begin{tabular}{p{0.2\linewidth}p{0.02\linewidth}p{0.75\linewidth}}
\raisebox{-2mm}{Rendered \LaTeX{}} & & Suppose $ { s_{8} } $ is convergent and $ { s_{7} } $ is convergent . 
Then $ \mathop { \rm lim } ( { s_{ 8 } } { + } { s_{7} } ) \mathrel { = } 
\mathop { \rm lim } { s_{8}} { + } \mathop { \rm lim } { s_{7} } $  \\
\raisebox{6mm}{Input \LaTeX{}} & &
\begin{BVerbatim}
Suppose $ { s _ { 8 } } $ is convergent and $ { s _ { 7 } } 
$ is convergent . Then $ \mathop { \rm lim } ( { s _ { 8 } 
} { + } { s _ { 7 } } ) \mathrel { = } \mathop { \rm lim } 
{ s _ { 8 } } { + } \mathop { \rm lim } { s  _ { 7 } } $ .
\end{BVerbatim}
\\
\raisebox{2mm}{Correct} & &
\begin{BVerbatim}
seq1 is convergent & seq2 is convergent implies lim ( seq1
+ seq2 ) = ( lim seq1 ) + ( lim seq2 ) ;
\end{BVerbatim}
\\
\raisebox{2mm}{Snapshot-1000} & &
\begin{BVerbatim}
x in dom f implies ( x * y ) * ( f | ( x | ( y | ( y | y )
) ) ) = ( x | ( y | ( y | ( y | y ) ) ) ) ) ;
\end{BVerbatim}
\\[-3mm]
Snapshot-2000 & &
\verb!seq is summable implies seq is summable ;!
\\[-3mm]
Snapshot-3000 & &
\verb!seq is convergent & lim seq = 0c implies seq = seq ;!
\\
\raisebox{2mm}{Snapshot-4000} & &
\begin{BVerbatim}
seq is convergent & lim seq = lim seq implies seq1 + seq2
is convergent ;
\end{BVerbatim}
\\
\raisebox{2mm}{Snapshot-5000} & &
\begin{BVerbatim}
seq1 is convergent & lim seq2 = lim seq2 implies lim_inf 
seq1 = lim_inf seq2 ;
\end{BVerbatim}
\\
\raisebox{2mm}{Snapshot-6000} & &
\begin{BVerbatim}
seq is convergent & lim seq = lim seq implies seq1 + seq2
is convergent ;
\end{BVerbatim}
\\
\raisebox{2mm}{Snapshot-7000} & &
\begin{BVerbatim}
seq is convergent & seq9 is convergent implies 
lim ( seq + seq9 ) = ( lim seq ) + ( lim seq9 ) ;
\end{BVerbatim}
\\
\raisebox{2mm}{Snapshot-8000} & &
\begin{BVerbatim}
seq1 is convergent & seq2 is convergent implies
lim seq1 = lim seq2 + lim seq2 ;
\end{BVerbatim}
\\
\raisebox{2mm}{Snapshot-9000} & &
\begin{BVerbatim}
seq1 is convergent & lim seq1 = lim seq2 implies ( seq1 
+ seq2 ) + ( lim seq1 ) = ( lim seq1 ) + ( lim seq2 ) ;
\end{BVerbatim}
\\
\raisebox{2mm}{Snapshot-10000} & &
\begin{BVerbatim}
seq1 is convergent & lim seq1 = lim seq2 implies
seq1 + seq2 is convergent ;
\end{BVerbatim}
\\
\raisebox{2mm}{Snapshot-11000} & &
\begin{BVerbatim}
seq1 is convergent & lim seq = lim seq1 implies
lim_sup seq1 + lim_sup seq2 = lim seq1 + lim seq2 ;
\end{BVerbatim}
\\
\raisebox{2mm}{Snapshot-12000} & &
\begin{BVerbatim}
seq1 is convergent & seq2 is convergent implies
lim ( seq1 + seq2 ) = ( lim seq1 ) + ( lim seq2 ) ;
\end{BVerbatim}
\\
\end{tabular}
\end{table}

\end{document}